\documentclass[sigconf, 9pt]{acmart}

\usepackage{booktabs}

\usepackage{boldline}

\usepackage{multirow}

\usepackage[]{graphicx}

\usepackage{footnote}
\makesavenoteenv{tabular}
\makesavenoteenv{table}

\usepackage[export]{adjustbox}
\usepackage[caption=false,font=footnotesize]{subfig} %

\usepackage{xcolor}
\colorlet{darkblue}{blue!40!black}
\colorlet{darkred}{red!50!black}

\usepackage[hidelinks,bookmarks=true,bookmarksnumbered=true,pdfusetitle,pdfdisplaydoctitle,unicode]{hyperref}
\hypersetup{colorlinks,%
    citecolor=ForestGreen,%
    filecolor=Orange,%
    linkcolor=blue,%
    urlcolor=BrickRed,
}

\usepackage{amsmath}
\usepackage{amsfonts}
\usepackage{mathtools}
\DeclarePairedDelimiter\ceil{\lceil}{\rceil}
\DeclarePairedDelimiter\floor{\lfloor}{\rfloor}

\usepackage{fixmath}

\usepackage{balance}

\usepackage{siunitx}

\usepackage{enumitem}

\usepackage{soul}

\usepackage{todonotes}

\usepackage{flafter}

\usepackage{algorithm}
\usepackage{algpseudocode}

\usepackage{url}

\usepackage{pifont}
\newcommand{\cmark}{\ding{51}}
\newcommand{\xmark}{\ding{55}}

\algnewcommand\algorithmicforeach{\textbf{for each}}
\algdef{S}[FOR]{ForEach}[1]{\algorithmicforeach\ #1\ \algorithmicdo}

\setcopyright{none}

\begin{document}
\renewcommand\footnotetextcopyrightpermission[1]{} %
\pagestyle{plain} %
\pagenumbering{gobble} %

\title{SynergicLearning: Neural Network-Based Feature Extraction for Highly-Accurate Hyperdimensional Learning}

\author{Mahdi Nazemi}
\affiliation{%
 \institution{University of Southern California}
}
\email{mnazemi@usc.edu}

\author{Amirhossein Esmaili}
\affiliation{%
 \institution{University of Southern California}
}
\email{esmailid@usc.edu}

\author{Arash Fayyazi}
\affiliation{%
 \institution{University of Southern California}
}
\email{fayyazi@usc.edu}

\author{Massoud Pedram}
\affiliation{%
 \institution{University of Southern California}
}
\email{pedram@usc.edu}

\begin{abstract}
    
Machine learning models differ in terms of accuracy, computational/memory complexity, training time, and adaptability among other characteristics. 
For example, neural networks (NNs) are well-known for their high accuracy due to the quality of their automatic feature extraction while brain-inspired hyperdimensional (HD) learning models are famous for their quick training, computational efficiency, and adaptability. 
This work presents a hybrid, synergic machine learning model that excels at all the said characteristics and is suitable for incremental, on-line learning on a chip. 
The proposed model comprises an NN and a classifier. 
The NN acts as a feature extractor and is specifically trained to work well with the classifier that employs the HD computing framework. 
This work also presents a parameterized hardware implementation of the said feature extraction and classification components while introducing a compiler that maps any arbitrary NN and/or classifier to the aforementioned hardware. 
The proposed hybrid machine learning model has the same level of accuracy (i.e. \(\pm\)1\%) as NNs while achieving at least 10\% improvement in accuracy compared to HD learning models. 
Additionally, the end-to-end hardware realization of the hybrid model improves power efficiency by 1.60x compared to state-of-the-art, high-performance HD learning implementations while improving latency by 2.13x. 
These results have profound implications for the application of such synergic models in challenging cognitive tasks. 
    
\end{abstract}

\maketitle

\section{Introduction}
\label{sec:intro}

Machine learning models have proven successful in solving a wide variety of challenging problems such as computer vision and speech recognition. 
They are commonly characterized by their level of accuracy, computational/memory complexity, training time, and adaptability  among other features. 
One can categorize machine learning models according to the aforesaid characteristics. 
For example, neural networks (NNs) typically achieve high accuracy \cite{he2016deep}, are computationally expensive \cite{sze2017efficient}, have long training times \cite{livni2014computational}, and tend to forget previously learned information upon learning new information (aka catastrophic forgetting) \cite{mccloskey1989catastrophic,sahoo2017online,losing2018incremental}. 
A machine learning model is more viable for on-chip learning (also called learning on-a-chip which refers to designing a custom chip that can be used for both training and inference) when it has low computational/memory complexity and supports one-pass training/fine-tuning while maintaining a high level of accuracy.

The main reason behind the high accuracy of NNs is their ability to automatically extract high-quality, high-level features from labeled data. 
AlexNet \cite{krizhevsky2012imagenet} is an outstanding example that clearly demonstrates the gap between the quality of features extracted by NNs compared to handcrafted features extracted by experts in the domain (in the ImageNet Large Scale Visual Recognition Challenge \cite{russakovsky2015imagenet}, AlexNet was able to achieve 10.8\% higher accuracy compared to the runner up, which used handcrafted features). 
Unfortunately, the high accuracy of NNs is accompanied by an enormous computational/memory cost during training and inference. 
Training an NN is a time-consuming, iterative process where in each iteration, all training data is applied to the model and the parameters of the model are updated according to stochastic gradient descent. 

As another example, hyperdimensional (HD) learning models train quickly, are highly adaptable and computationally efficient (compared to NNs), but suffer from lower levels of accuracy compared to NNs \cite{kanerva2009hyperdimensional}. 
HD learning uses randomly generated, high-dimensional vectors to project training data into HD space such that samples belonging to the same class are placed in close proximity of each other, forming a cluster in the HD space. 
It then defines HD centroids that represent different classes. 
This relatively simple training process only requires one pass over the training data. 
It also enables efficient incremental, lifelong learning because updating the model with new training data is as simple as updating the cluster centroids. 
The major disadvantage of HD learning is that it works with raw or handcrafted input features, which are inferior to the ones extracted by NNs. 

The complementary characteristics of NNs and HD models encourage the introduction of a hybrid, synergic machine learning model that builds on their strengths while avoiding their shortcomings. 
However, simply employing NNs for feature extraction and HD models for classification so as to enable on-chip learning has the following challenges. 
Not only is the training of NNs for feature extraction an iterative, energy-consuming process but also it requires access to both previous training data and newly provided data to avoid catastrophic forgetting. 
Therefore, frequent weight updates of NNs can be extremely costly in the context of learning on-a-chip. 
Additionally, the HD learning models that work well for solving cognitive tasks have a huge number of dimensions, e.g., 10,000, which requires their hardware implementation to time-share resources and therefore, have a relatively high latency. 
This prevents real-time fine-tuning of the model when new training data becomes available. 
Moreover, training NNs for feature extraction separately from the design of the HD learning model produces suboptimal results because it does not account for the effect of HD classification layers on the NN feature extraction layers and vice versa. 
This means that the prediction/classification accuracy of the overall hybrid solution will suffer. 

This work presents SynergicLearning, a hybrid learning framework for incremental, on-line learning on a chip. 
SynergicLearning is comprised of three components which enable end-to-end learning:
\begin{enumerate}
    \item{\textbf{A Two-step Training Approach:} This training approach first trains an NN while including some components of the HD learning system in the NN's training loop to learn high-quality, high-level features that are specifically tailored for the HD learning system. It then passes training data (including the initial data as well as the ones that are generated during the lifetime of the model) through the feature extraction layers of the NN to provide features for training/fine-tuning of the HD classifier (the neural network parameters are fixed at this step). Such a two-level training approach enables automatic feature extraction while reducing the number of dimensions in the HD classifier by two to three orders of magnitude\footnote{While the term \textit{hyperdimensional} learning is no longer applicable to such a classifier, we keep using the same term to highlight the fact that the operations used in the classifier are based on those defined in the hyperdimensional computing framework.}.}
    \item{\textbf{An On-chip Learning Module:} This module is comprised of parameterized NN and HD processing modules, which respectively execute operations required by the NN feature extraction layers and operations required by the HD classifier. The NN processing module includes a systolic array which performs vector-matrix multiplications and an ALU which supports operations such as batch normalization, pooling, and ReLU. The HD processing module supports the arithmetic operations defined in the HD computing including \textit{binding}, \textit{bundling}, and distance calculation (Section~\ref{sec:prelim} details these operations). The parameterized hardware implementation enables efficient exploration of the design space to find configurations that satisfy the design constraints such as energy and resource utilization.}
    \item{\textbf{A Compiler:} The custom compiler performs code optimizations and generates instructions that efficiently schedule different operations required by the NN feature extraction and HD classification steps (e.g., vector-matrix multiplications and data movement) on the target platform.}
\end{enumerate}
Table~\ref{table:nn-hd-nnhdl} compares different characteristics of NNs, HD learning systems (HDL), and the proposed SynergicLearning approach. 
\begin{table*}[tb]
    \centering
    \caption{Comparison of different characteristics of NNs, HD learning systems, and SynergicLearning.}
    \begin{tabular*}{0.8\textwidth}{l @{\extracolsep{\fill}} *{4}{c}}
        \toprule
        Machine Learning        & Automatic             & High      & One-pass              & Adaptable w/o Accessing   \\
        Model                   & Feature Extraction    & Accuracy  & Training/Fine-tuning  & Previous Training Samples \\
        \hlineB{3}
        NN                      & \cmark                & \cmark    & \xmark                & \xmark                    \\
        HDL                     & \xmark                & \xmark    & \cmark                & \cmark                    \\
        SynergicLearning        & \cmark                & \cmark    & \cmark                & \cmark                    \\
        \bottomrule
    \end{tabular*}
    \label{table:nn-hd-nnhdl}
\end{table*}
It is observed that SynergicLearning enjoys automatic feature extraction and high accuracy because it employs an NN that is tailored for HDL. 
Furthermore, it only requires one pass to train/fine-tune its HD classifier and last but not least, it does not require accessing previous training samples to update the model when new data becomes available. 

The remainder of this paper is organized as follows. 
Section~\ref{sec:prelim} explains the preliminaries on HD computing, discusses some of its shortcomings, and motivates the presented solution. 
Next, Section~\ref{sec:proposed-method} details the proposed learning framework while Section~\ref{sec:hw-architecture} explains the proposed hardware architecture and compiler for inference.
After that, Section~\ref{sec:results} presents the experimental results while Section~\ref{sec:related} briefly reviews the related work on HD computing. 
Finally, Section~\ref{sec:conclusions} concludes the paper.

\section{Preliminaries \& Motivation}
\label{sec:prelim}

HD computing defines a new computation framework that relies on high-dimensional random vectors (aka \textit{hypervectors}) and the arithmetic operations that manipulate such large random patterns. 
An HD system starts by randomly generating \(d^h\)-dimensional, holistic seed hypervectors with independent and identically distributed (i.i.d) elements.
This means that the information encoded into each hypervector is uniformly distributed over all its elements. 
Therefore, unlike the conventional computing framework, elements in different bit positions in hypervectors are equally significant. 
The seed hypervectors are typically stored in a memory called the \textit{cleanup memory}. 
The arithmetic operations defined on the seed hypervectors, e.g. binding and bundling, enable meaningful computations in the corresponding hyperspace. 
The focus of this paper is on binary hypervectors where each element is equally likely to be a zero or one. 
Binary hypervectors enjoy simplified, hardware-friendly arithmetic operations. 

The distance between two binary hypervectors is measured in normalized Hamming distance, i.e. the number of bit positions where the values of hypervectors differ, divided by \(d^h\). 
Consequently, the distance is always in the range zero to one inclusive. 
However, because the distance between two randomly generated hypervectors follows a binomial distribution, most hypervectors are about 0.5 apart from one another (when \(d^h\) is large) and therefore, are nearly orthogonal (aka \textit{unrelated}). 
Additionally, flipping the values of a relatively large portion of elements in a hypervector, e.g. one-third of all elements, results in a hypervector that is closer to the original hypervector compared to its unrelated hypervectors. 
This results in considerable tolerance to noise and approximation.
When the cleanup memory is queried with a noisy hypervector, it returns the seed hypervector that is closest to the input query, hence the name cleanup.

Two of the commonly used arithmetic operations in HD computing are binding and bundling. 
The binding operation is used for variable-value association. 
Assume variable \(z\) and its corresponding value \(z_0\) are represented with unrelated hypervectors \(\mathbold{z}\) and \(\mathbold{z}_0\), respectively. 
Then, the bound pair \(z = z_0\) can be represented by \(\mathbold{z} * \mathbold{z}_0\), where element-wise multiplication (\(*\)) is replaced with element-wise XOR for binary hepervectors. 
The resulting hypervector is unrelated to both \(\mathbold{z}\) and \(\mathbold{z}_0\). 
However, each original hypervector can be recovered from the resulting hypervector given the other, e.g. \(\mathbold{z}_0 = S((\mathbold{z} * \mathbold{z}_0) * \mathbold{z})\), where \(S(.)\) looks up the cleanup memory. 
This process is called \textit{unbinding}. 
The bundling operation condenses a list of hypervectors into a single representative hypervector that is similar to all its constituents. 
This is achieved by summing up all hypervectors, followed by the comparison of each element in the resulting (summation) hypervector with half the number of original hypervectors to create a binary hypervector. 
If the original hypervectors are bound, their variables and/or values can be found through unbinding the bundled hypervector. 

The HD computing framework can be used to solve cognitive tasks such as speech recognition and activity recognition \cite{kanerva2009hyperdimensional,imani2018hierarchical,morris2019comphd}. 
Alg.~\ref{alg:hdl} summarizes different steps of training an HD model. 
The inputs to the algorithm are \(d^l\)-dimensional input features, their corresponding labels/classes, the dimension of hypervectors (\(d^h\)), and the number of quantization levels used to discretize the input values while the outputs are the HD centroids representing each class. 
The training starts with the generation of seed hypervectors for all \(d^l\) features as well as the quantized values they can assume. 
While the seed hypervectors for features are generated randomly, the ones for quantized values are found by randomly flipping a specific number of bits of a seed hypervector to ensure the similarity of the hypervectors representing nearby values. 
Next, each feature and its value are bound and the set of all bound hypervectors are bundled into a single hypervector (aka \textit{encoding}). 
Finally, the encoded hypervectors are categorized according to their labels and the set of hypervectors belonging to a class are bundled to find a representative centroid. 
\begin{algorithm}[tb]
	\caption{Training an HD Model}
    \label{alg:hdl}
		
	\begin{algorithmic}[1]
		\Require
        	\Statex \(\mathbold{X}_{n \times d^l} = \mathbold{x}_{1..n}, \mathbold{x}_i \in \mathbb{R}^{d^l}\) \Comment{the low-dimensional input features}
        	\Statex \(\mathbold{y} = y_{1..n}, 1 \leq y_i \leq c\) \Comment{the target labels/classes}
            \Statex \(d^h\) \Comment{the number of hyperspace dimensions}
			\Statex \(q\) \Comment{the number of quantization levels}
   		\Ensure
   			\Statex \(\mathbold{T}_{c \times d^h} = \mathbold{t}_{1..c}\) \Comment{the HD centroids}
			
		\State generate \(\mathbold{S}_{d^l \times d^h} = \mathbold{s}_{1..d^l}\) \Comment{seed hypervectors for features}
        \State \(p = \floor{\frac{d^h}{q}}\) \Comment{number of bits to flip}
        \State generate \(\mathbold{q}_1\) randomly
        \For{\(i = 2..q\)}
            \State \(\mathbold{q}_i = \text{randomly pick } p \text{ unflipped bits and flip them in } \mathbold{q}_{i - 1}\)
        \EndFor
		\State \(\mathbold{Q}_{q \times d^h} = \mathbold{q}_{1..q}\) \Comment{seed hypervectors for levels}
		
		\ForEach{\(\mathbold{x}_i\)} \Comment{\textbf{encode all samples}}
		    \State \(\mathbold{x}^q_i = \mathrm{quantize}(\mathbold{x}_i, q)\) \Comment{quantize real values to integers}
		    \State \(X_i = \emptyset\)
		    \For {\(j\) in \(1..d^l\)} \Comment{\textbf{bind feature-value pairs}}
		        \State \(X_i = X_i \cup \mathrm{bind}(\mathbold{s}_j, \mathbold{q}_{x^q_{i_j}})\)
		    \EndFor
		    \State \(\mathbold{x}^{enc}_i = \mathrm{bundle}(X_i)\) \Comment{\textbf{bundle bound hypervectors}}
		\EndFor
		
		\State \(T_1 = T_2 = ... = T_c = \emptyset\)
		\ForEach{\(\mathbold{x}^{enc}_i\)} \Comment{\textbf{group encoded inputs by labels}}
	        \State \(T_{y_i} = T_{y_i} \cup \mathbold{x}^{enc}_i\)
		\EndFor
		
		\ForEach{\(T_k\)} \Comment{\textbf{bundle all members of each class}}
	        \State \(\mathbold{t}_k = \mathrm{bundle}(T_k)\)
		\EndFor		
		
   		\State return \(\mathbold{T}\)
	\end{algorithmic}
\end{algorithm}
During inference, the closest centroid to an encoded test sample (in terms of normalized Hamming distance) determines the model's prediction. 

\begin{figure}[tb]
    \centering
    \includegraphics[width=0.9\columnwidth]{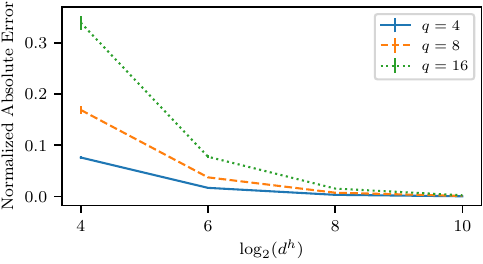}
    \caption{The mean and standard deviation of the normalized absolute error between the input features and the decoded features of their encoded hypervectors for different values of \(d^h\) and \(q\). Ideally, this error should be zero everywhere. However, the error has a non-zero value even at extremely high dimensions (\(d^h \simeq 10,000\)).}
    \label{fig:reconstruction-error}
\end{figure}

By fixing \(d^h\) and increasing \(q\), the hypervectors representing different quantization levels become more similar because fewer bits are flipped across consecutive hypervectors. 
This, in turn, complicates the unbinding process because the cleanup memory may return the wrong values. 
Fig.~\ref{fig:reconstruction-error} clearly illustrates this phenomenon by depicting the mean and standard deviation of the normalized absolute error between the input features and the decoded features of their encoded hypervectors. 
Ideally, decoding encoded hypervectors should return the exact same low-dimensional features as the original inputs (i.e., zero error), but this does not happen in practice. 
We believe this phenomenon is the main reason for the relatively poor performance of HD models compared to some other machine learning models such as NNs. 
\textbf{Therefore, creating input features that are aware of the error due to very similar quantization levels can improve classification accuracy significantly, especially at lower \(\boldsymbol{d^h}\).} 

\section{Proposed Method}
\label{sec:proposed-method}

\begin{figure*}[tb]
    \centering
    \includegraphics[width=0.9\textwidth]{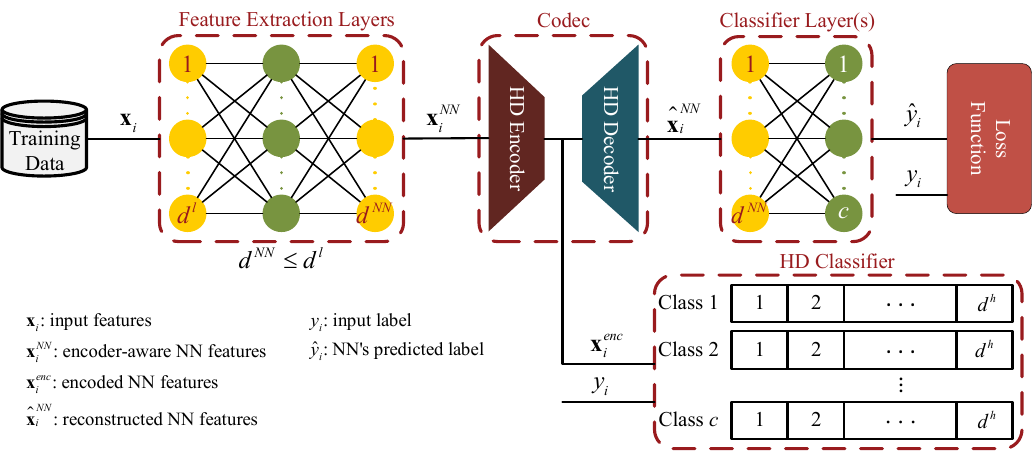}
    \caption{A high-level overview of the SynergicLearning framework. First, an encoder-aware NN is trained to extract high-quality, high-level features (top row of the figure). Next, encoded NN features are provided to train an HD classifier. Finally, during inference, the feature extraction layers of the NN and the HD classifier are both utilized to predict each test sample's label.}
    \label{fig:nnhdl}
\end{figure*}

Fig.~\ref{fig:nnhdl} demonstrates a high-level overview of the proposed hybrid learning framework. 
The proposed framework comprises two major components: an encoder-aware NN for high-quality feature extraction and an HD classifier. 

The NN includes feature extraction layers, an HD encoder-decoder pair (i.e. HD codec), and classifier layer(s). 
It takes the input features, passes them through the said components (aka forward propagation), and calculates a loss value by comparing the predicted labels with the expected ones. 
It then updates the model parameters, i.e. weights and biases, by backpropagating the loss value using the derivative of the operations defined in the NN. 
Because the operations defined in the codec are not differentiable and the fact that an ideal codec should behave like the identity function, the codec's derivative is approximated with that of the identity function during backpropagation. 

Pre-processing input features with an NN has numerous important advantages. 
First, \textbf{including the codec in the training loop encourages the NN to adjust its parameters such that it minimizes the impact of the codec's error on classification accuracy.} 
Training two identically initialized NNs, one including a codec and the other without a codec, would result in a completely different set of parameters. 
Second, the number of features extracted by the NN (\(d^{NN}\)) can be much lower than the number of low-dimensional input features (\(d^l\)), which in turn reduces the complexity of the HD classifier. 
Third, because the NN extracts encoder-aware features, \(d^h\) can be reduced by two to three orders of magnitude compared to the existing HD systems. 
In other words, because the degree of similarity of hypervectors representing quantization levels is less concerning, lower \(d^h\) values can work equally well. 
Fourth, there is a large body of work on reducing the complexity of NNs through quantization \cite{choi2018pact,zhou2016dorefa}, pruning \cite{zhang2018systematic}, and knowledge distillation \cite{hinton2015distilling}, to name but a few. 
This allows training NNs that are lightweight, thereby adding little overhead to the overall hardware cost. 

The HD classifier is very similar to the one described in Alg.~\ref{alg:hdl}. 
The only difference is that it takes the output of NN's feature extraction layers (\(\mathbold{x}^{NN}_i\)) instead of the original input features (\(\mathbold{x}_i\)). 
Therefore, it not only benefits from the inherent strength of NNs in feature extraction but also enjoys features that are specifically tailored for the HD encoder.

\section{Proposed Hardware Architecture \& Compiler}
\label{sec:hw-architecture}

The proposed hardware architecture implements an end-to-end, fully-parameterized implementation of SynergicLearning for inference. 
It consists of two major hardware components: an NN processing module, which includes a systolic array and an ALU, and a fully-parallel HD processing module which supports various operations such as binding, bundling, and distance calculation. 

\subsection{NN Processing Module}
\label{sec:NN-module}

Fig.~\ref{fig:NN-arch} demonstrates a high-level overview of the NN processing module, which comprises the following components: 
\begin{figure}[tb]
    \centering
    \includegraphics[width=\columnwidth]{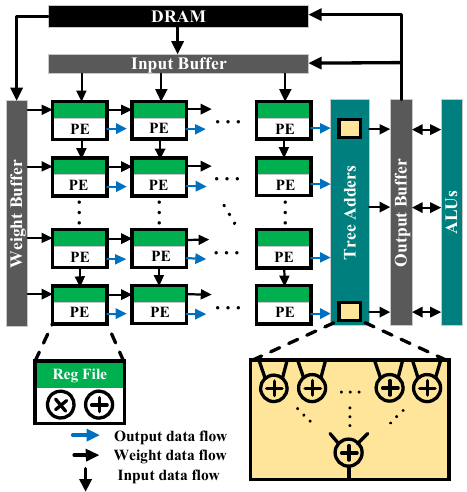}
    \caption{Architectural view of the NN processing module which includes a systolic array, on-chip memories, tree adders, and ALUs.}
    \label{fig:NN-arch}
\end{figure}
\begin{itemize}
    \item the systolic array, which consists of a two-dimensional array of processing elements, 
    \item on-chip memories (i.e. weight, input, and output buffers), which act as an intermediate storage between the DRAM and the systolic array, 
    \item tree adders, each of which performs a summation over a row of the systolic array, and
    \item ALUs, which support activation functions, batch normalization, pooling, etc.
\end{itemize}
Processing each layer of the NN requires the following operations. 
First, the weights are read from the external memory (DRAM) and stored in the weight buffer while inputs are read either from the DRAM or the output buffer. 
Next, the systolic array and tree adders calculate the neurons' pre-activation values by implementing vector-matrix multiplications. 
Then, ALUs apply batch normalization, activation function, pooling, etc. to pre-activations to generate the output features.
Finally, the output features are either rerouted to the input buffers or written back to the DRAM. 

The systolic array implements a weight-stationary dataflow \cite{sze2017efficient}, which reuses each weight value in different computations involved in vector-matrix multiplication and therefore, reduces the overhead associated with data movement. 
In this dataflow, the number of cycles it takes to process each layer of the NN is approximated by
\begin{equation*}
\label{eq:syslat}
    (\ceil[\big]{\frac{d_{l_i}}{w_{sys}}} + \log_2 w_{sys}) \times \ceil[\big]{\frac{d_{l_{i+1}}}{h_{sys}}}\mathrm{,}
\end{equation*}
where \(d_{l_i}\) is number of neurons in the \(i^\mathrm{th}\) layer and \(w_{sys}\) (\(h_{sys}\)) is the number of columns (rows) in the systolic array. 
\(\log_2 w_{sys}\) represents the depth of each tree adder. 

For mapping a neural network presented in high-level languages to the target FPGA, we developed an in-house compiler called SynergicCompiler. 
Since all hardware designs presented in this paper perform the same computation, i.e. a three-level nested loop for fully connected layers, the space explorations are defined by transformations (e.g., block, reorder, and parallelize) on the nested loop. 
Therefore, the compiler tries various choices of loop ordering and hardware parallelism for computing these nested loops of NNs and finds the most efficient one in terms of latency. 
The SynergicCompiler also generates a static schedule for the data movements between hierarchies of memories, e.g., between external memories and buffers, and buffers and registers within PEs. 
Static scheduling mitigates the need for complex handshaking and improves the scalability and performance of the processing modules. 
Finally, the compiler delivers a set of instructions that efficiently schedule different operations such as vector-matrix multiplications and data movement on the target platform. 
More details about the compiler are not included in this paper for brevity.

\subsection{HD Processing Module}
\label{sec:HD-module}

Fig.~\ref{fig:HD-arch} demonstrates a high-level overview of the pipelined HD processing module, which comprises the following components: 
\begin{figure*}[tb]
    \centering
    \includegraphics[width=\textwidth]{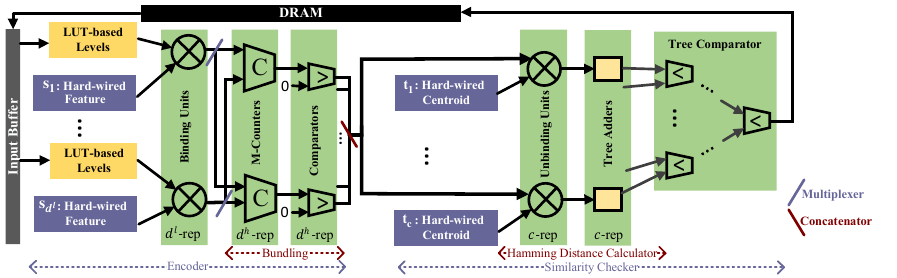}
    \caption{Architectural overview of the HD processing module which includes lookup tables that store hypervectors representing quantized levels, binding/unbinding units, majority counters, comparators, tree adders, and tree comparators.}
    \label{fig:HD-arch}
\end{figure*}
\begin{itemize}
    \item lookup tables (LUTs) that store hypervectors representing quantized levels, 
    \item binding/unbinding units, which perform parallel XOR operations,
    \item majority counters, which compute the population count of a bit vector by incrementing a \((\log_2 (d^l + 1) + 1)\)-bit counter when a set bit is encountered and decrementing it when a reset bit is seen,
    \item comparators, which produce binary hypervectors from integer hypervectors, and 
    \item tree adders and tree comparators, which implement a fully-parallel Hamming distance calculation and therefore, produce outputs in constant time. 
\end{itemize}

The architecture of the proposed HD processing module has the lowest achievable latency but suffers from high resource consumption at large \(d^h\) values compared to other possible architectures such as the ones explained in \cite{schmuck2019hardware}. 
However, because SynergicLearning allows the utilization of HD learning systems with extremely low \(d^h\) values, the resource usage of the HD processing module will be negligible. 
Furthermore, because all the aforementioned components produce their results in constant time, the final output of the HD processing module will be produced in constant time too. 
Additionally, because of the pipelined implementation of the HD processing module, it can produce an output every cycle, hence very high throughput.

\section{Results \& Discussion}
\label{sec:results}

\subsection{Experimental Setup}

\subsubsection{\textbf{Datasets}}

To study the effectiveness of SynergicLearning, we use two publicly available datasets: Human Activity Recognition (HAR) \cite{anguita2013public} and ISOLET \cite{cole1990isolet}. 
HAR includes 10,299 samples, each of which contains 561 handcrafted features and a label that corresponds to one of six possible activities. 
ISOLET, on the other hand, contains 7,797 samples, each of which includes 617 handcrafted features and a label that corresponds to one of the 26 characters in the English alphabet. 
The goal is to take the input features and their labels and train classifiers that predict labels of unseen samples accurately.

\subsubsection{\textbf{Training Framework}}

We implement a PyTorch-compatible \cite{paszke2019pytorch} HD computing library that includes operations such as binding/unbinding, bundling, encoding, and decoding. 
Because of the compatibility with PyTorch, the operations can be mapped efficiently to either CPUs or GPUs. 
Additionally, they can be easily integrated into existing PyTorch designs such as NNs. 

We also implement a training ecosystem that takes a user-defined (possibly existing) NN architecture and the parameters of the HD learning system (e.g. \(d^h\) and \(q\)) and automatically glues different components together to enable encoder-aware training of the neural network. 
Similarly, it includes easy-to-use HD training modules. 
This training ecosystem allows us to quickly explore different designs and compare their accuracy.

\subsubsection{\textbf{Neural Network Training}}

We train all NNs by minimizing a cross-entropy loss function for 120 epochs, with a batch size of 256, and an \(l_2\) regularizer. 
Additionally, we use a learning rate scheduler similar to the one described in \cite{smith2019super} where the maximum learning rate is set to 0.01 while the number of steps per epoch is 25.

\subsubsection{\textbf{Hardware Emulation Framework}}

To implement the NN and HD processing modules, we use the Xilinx SDAccel which provides a toolchain for programming and optimizing different applications on Xilinx FPGAs using a high-level language (C, C++ or OpenCL) and/or hardware description languages (VHDL, Verilog and SystemVerilog), as well as a runtime based on the OpenCL APIs that can be used by the host-side software to interact with the accelerator. 
We evaluate our proposed architecture using SDAccel on the ISOLET dataset targeting the Xilinx UltraScale+ VU9P FPGA on AWS EC2 F1 instances. 
We also use the Vivado power report provided by Xilinx to assess the power consumption of each design.

\subsection{The Impact of NNs on the Quality of HD Features}

In this section, we study the impact of NNs on the quality of encoded HD features by visualizing different samples of the HAR dataset in two-dimensional (2D) space. 
The feature extraction layers of the NNs consist of two fully-connected layers, each of which has 561 neurons. 
We deliberately keep the number of neurons in the final feature extraction layer the same as the one for the input features (i.e. \(d^{NN} = d^l\)) to ensure the difference across the results of various experiments is only due to the introduction of NNs. 
We use ReLU and PACT \cite{choi2018pact} for the activation functions of the first and second layer, respectively.

\begin{figure*}[tb]
    \centering
    \subfloat[HDL] {
		\includegraphics[width=0.3\textwidth, valign=c]{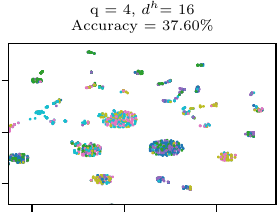}
		\label{fig:11}
	}
    \subfloat[NN followed by HDL] {
		\includegraphics[width=0.3\textwidth, valign=c]{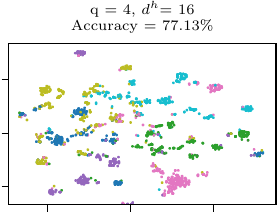}
		\label{fig:112}
	}
    \subfloat[Encoder-aware NN followed by HDL] {
		\includegraphics[width=0.3\textwidth, valign=c]{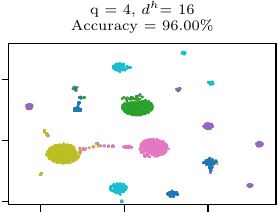}
		\label{fig:211}
	}
	
    \subfloat[HDL] {
		\includegraphics[width=0.3\textwidth, valign=c]{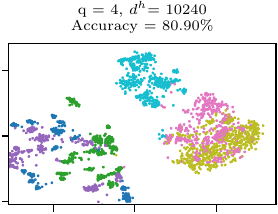}
		\label{fig:15}
	}
    \subfloat[NN followed by HDL] {
		\includegraphics[width=0.3\textwidth, valign=c]{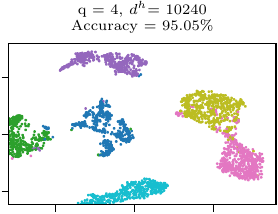}
		\label{fig:116}
	}
    \subfloat[Encoder-aware NN followed by HDL] {
		\includegraphics[width=0.3\textwidth, valign=c]{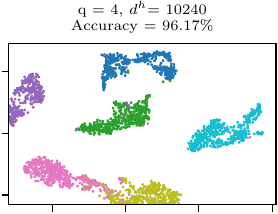}
		\label{fig:215}
	}
	
	\caption{Two-dimensional (t-SNE) representation of the encoded hypervectors of the HAR dataset for three different designs: HDL, NN followed by HDL, and encoder-aware NN followed by HDL.}
	\label{fig:TSNE}
\end{figure*}

Fig.~\ref{fig:TSNE} shows the 2D representation of the encoded hypervectors of the test set for three different designs: HDL, NN followed by HDL, and encoder-aware NN followed by HDL (i.e. the proposed flow). 
To obtain the 2D representation, we employ t-distributed stochastic neighbor embedding (t-SNE) \cite{maaten2008visualizing}, which is a technique used for visualizing high-dimensional data. 
t-SNE tends to provide good visualizations because it tries to keep the similarities in HD space in the 2D representation as well. 
The 2D representations of hypervectors belonging to different classes are shown using different colors. 
For figures \ref{fig:11}-\ref{fig:211}, we use \(d^h=16\), and for figures \ref{fig:15}-\ref{fig:215}, we use \(d^h=10,240\). 
For all experiments, \(q = 4\). 

For small values of \(d^h\) (e.g. 16), it is observed that HDL performs poorly in the separation of points in the HD space (Fig.~\ref{fig:11}). 
On the other hand, the addition of an NN to the flow helps with more proper separation of data points (Fig.~\ref{fig:112}) while introducing an encoder-aware NN leads to a near-perfect clustering of data (Fig.~\ref{fig:211}). 
The accuracy values reported in Fig.~\ref{fig:11}-\ref{fig:211} further support this observation. 
For large values of \(d^h\) (e.g. 10,240), it is observed that HDL performs relatively well while models that include NNs still outperform the HDL model by a large margin (Fig.~\ref{fig:15}-\ref{fig:215}). 
In this configuration, the model that includes an NN and the one that has an encoder-aware NN perform almost equally well.

\subsection{Comparison of Classification Accuracy}

Table~\ref{table:acc-comp} compares the highest values of accuracy reported for NNs and HD learning systems with the proposed SynergicLearning approach on the HAR and ISOLET datasets. 
\begin{table}[t]
    \centering
    \caption{Top accuracy reported for NNs, HD learning systems, and SynergicLearning on HAR and ISOLET datasets.}
    \begin{tabular*}{0.4\textwidth}{c|cc}
        \toprule
        Dataset &  Machine Learning Model & Accuracy (\%)  \\
        \hlineB{3}
        \multirow{3}{*}{{\rotatebox[origin=c]{90}{\textbf{HAR}}}}  & NN \cite{ignatov2018real} \(^{\ddagger}\)\footnote{They also reported higher accuracy of 97.6 \% when they added statistical features and data centering methods to their convolutional neural network.} & 95.31 \% \\
        & HDL \cite{imani2019fach}  & 93.4\% \\
        & SynergicLearning & 96.44 \%  \\
        \hlineB{3}
        \multirow{3}{*}{{\rotatebox[origin=c]{90}{\textbf{ISOLET}}}}  & NN \cite{chechik2009online, imani2017voicehd}\(^{\mathrm{*}}\)  & 95.9 \%\\
        & HDL \cite{imani2017voicehd}   & 93.8 \% \\
        & SynergicLearning & 96.67 \%  \\
        \bottomrule
        \multicolumn{3}{l}{\(^{\ddagger}\)Uses a convolutional neural network.} \\
        \multicolumn{3}{l}{\(^{\mathrm{*}}\)Uses a fully-connected network with 48 hidden layers.}
    \end{tabular*}
    \label{table:acc-comp}
\end{table}
It is observed that on these datasets, the proposed hybrid model outperforms both NNs and HD learning systems used in the prior work. 

\begin{figure}[tb]
    \centering
    \subfloat {
		\includegraphics[width=0.48\columnwidth, valign=c]{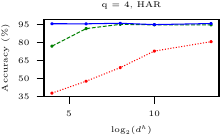}
	}
    \subfloat {
		\includegraphics[width=0.48\columnwidth, valign=c]{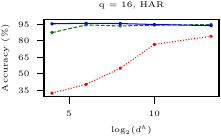}
	}
	
    \subfloat {
		\includegraphics[width=0.48\columnwidth, valign=c]{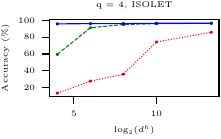}
	}
    \subfloat {
		\includegraphics[width=0.48\columnwidth, valign=c]{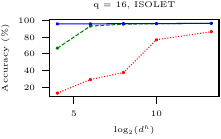}
	}
	
    \subfloat {
		\includegraphics[width=\columnwidth, valign=c]{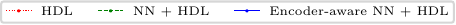}
	}
    \caption{Classification accuracy of different models on HAR and ISOLET datasets for different values of \(d^h\) and \(q\).}
    \label{fig:accuracy_har_isolet}
\end{figure}

Fig.~\ref{fig:accuracy_har_isolet} compares classification accuracy of three different models (HDL, NN followed by HDL, and encoder-aware NN followed by HDL) for different values of \(d^h\) and \(q\) on HAR and ISOLET datasets.%
It is observed that the model that includes an NN consistently outperforms HDL while the model that has an encoder-aware NN outperforms the other two in almost all experiments. 
On the HAR dataset, the difference between the model with an encoder-aware NN and the HDL model is as large as about 63\% at \(d^h = 16\) while it decreases to about 14\% at \(d^h = 10,240\). 
Similarly, On the ISOLET dataset, the difference between the model with an encoder-aware NN and the HDL model is as large as about 83\% at \(d^h = 16\) while it decreases to about 10\% at \(d^h = 10,240\). 

Another key observation is that the model with an encoder-aware NN achieves almost the same level of accuracy at different values of \(d^h\). 
This is particularly interesting from a hardware cost perspective, because we can pick the lowest value of \(d^h\) (16 in these experiments) and achieve significant reduction in resource utilization while maintaining high accuracy. 

We also study the effect of different random seeds for initialization of NN weights and randomly generated seed hypervectors on classification accuracy. 
Based on our experiments, the difference between the lowest and highest values of classification accuracy across designs that use different seeds is at most 1\%. 
We believe such variation in classification accuracy is acceptable.

\subsection{Incremental Learning}

Table~\ref{table:incremental} compares the accuracy of HD learning models and SynergicLearning when a portion of data is initially used for training while the remaining data is used for fine-tuning the model on a chip. 
Because on-chip-learning is extremely costly for NNs, we do not consider them in this comparison. 
\begin{table}[tb]
    \centering
    \caption{Comparison of the effect of incremental learning on the accuracy of different models on the ISOLET dataset.}
    \begin{tabular*}{0.43\textwidth}{ccccc}
        \toprule
        Machine Learning &  \multicolumn{4}{c}{Accuracy}  \\
        Model & \multicolumn{4}{c}{(Ratio of the Initial Training Data)} \\
        \hlineB{3}
        \multirow{2}{*}{HDL} & 85.76\%  & 85.76\% & 85.76\% & 85.76\%    \\
         & (0.25) & (0.5) & (0.75) & (1)    \\
        \multirow{2}{*}{SynergicLearning} & 86.21\%  & 91.21\%  & 94.03\% & 95.77\%   \\
        &(0.25)  & (0.5)   & (0.75) & (1) \\
        \bottomrule
    \end{tabular*}
    \label{table:incremental}
\end{table}
As expected, the HDL model is insensitive to whether the training data is provided incrementally or all at once and therefore, its accuracy remains constant and relatively low. 
For the SynergicLearning model, on the other hand, the accuracy keeps increasing when more data is provided to the NN in the initial training phase because it allows the NN to find higher quality features. 
\begin{table*}[tb]
  \caption{Comparison between the hardware metrics of SynergicLearning (\(d^h=16\)) with pure HD (\(d^h =10,240\)) over the ISOLET dataset on Xilinx UltraScale+ VU9P FPGA. The improvements of our approach compared to other approaches are shown in parantheses.} 
  \label{tab:hardware-results}
  \resizebox{\textwidth}{!}{%
  \centering %
  \begin{tabular}{c|c|cccccc} %
  \toprule[\heavyrulewidth]\toprule[\heavyrulewidth]
  \textbf{Approach} & \textbf{Implementation} &\textbf{BRAMs-18K (\%)} & \textbf{DSPs-48E (\%)} & \textbf{FFs (\%)} & \textbf{LUTs (\%)} & \textbf{Latency (\(\mu\)s)} & \textbf{Power (W)} \\
      \midrule
  \multirow{1}{*}[-0.4ex]{{\textbf{SynergicLearning}}}
    & \textbf{NN+HD}                 & 1.8     & 15.0      & 0.8          & 5.1                & 23.3          & 5.3 \\
    \midrule
  \multirow{2}{*}[-0.4ex]{{\textbf{Pure HD \cite{schmuck2019hardware}}}} 
  & \textbf{Parallel}               & 0 (N/A)  & 0 (N/A)   & 11.0 \textbf{(93\%)}  & 15.0 \textbf{(66\%)}        & 49.5 \textbf{(53\%)}   & 8.5 \textbf{(38\%)} \\
  \cmidrule{2-8}
  & \textbf{Sequential}             & 0 (N/A)  & 0 (N/A)   & 11.0 \textbf{(93\%)}  & 9.0 \textbf{(43\%)}         & 788.7 \textbf{(97\%)}     & 7.7 \textbf{(31\%)}\\
  \midrule
  \textbf{NN \cite{chechik2009online, imani2017voicehd}} & \textbf{Systolic Array} & 1.7 (-6\%) & 15.0 (0\%) & 0.7 (-14\%)  & 3.6 (-42\%)   & 835.9 \textbf{(97\%)} & 5.1 (-4\%)   \\
  \bottomrule[\heavyrulewidth] 
  \end{tabular}}
\end{table*}
This encourages less frequent, off-line updates to the NN for increasing the accuracy of the model. 

\subsection{The Hardware Cost of NN \& HD Processing Modules}

\begin{figure}[tb]
    \centering
	\includegraphics[width=0.7\columnwidth]{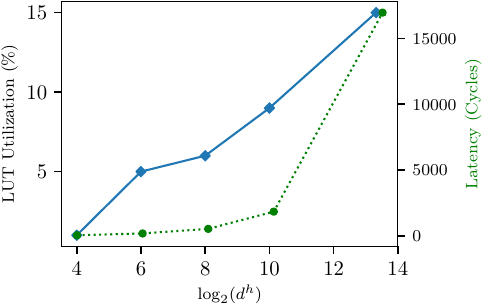}
    \caption{the LUT utilization and latency of HD processing modules for different values of \(d^h\).}
    \label{fig:LUT-latency-d}
\end{figure}

Fig.~\ref{fig:LUT-latency-d} shows the LUT utilization and latency of HD processing modules for different values of \(d^h\) while limiting the number of adders in each stage of tree adders to 16. 
It is observed that the latency grows very rapidly when increasing \(d^h\) to values required for meeting accuracy requirements. 
Additionally, to reduce the resource utilization for large values of \(d^h\), we can change the architecture from a fully-parallel architecture to a vector-sequential architecture where all adders and counters operate in a sequential manner (compare Sequential Implementation with Parallel Implementation entries in Table.~\ref{tab:hardware-results}). While our parameterized architecture has a capability to generate both parallel and sequential-vector for HD processing module of SynergicLearning approach, we report the results for parallel implementation which delivers higher performance. Thanks to exteremly low \(d^h\) value in SynergicHD, the hardware overhead of parallel implementation is minimal.

Table~\ref{tab:hardware-results} compares area utilization, latency, and power consumption of SynergicLearning at \(d^h=16\) with pure HD processing module at \(d^h=10,240\). 
SynergicLearning outperforms the fully-parallel pure HD processing module in terms of latency by a factor of 2.13x while yielding 1.60x lower power consumption. 
Compared to the vector-sequential implementation of HD processing module, SynergicLearning achieves 33.89x improvement in latency while yielding 1.45x lower power consumption. 

It is worth mentioning that our designs are capable of achieving high clock rates (i.e. 344 MHz). 
The breakdown of different metrics between the NN and HD processing modules is as follows. 
The NN processing module consumes 93\%, 100\%, 87\%, and 71\% of the total consumed BRAMs-18K, DSPs-48E, FFs, and LUTs, respectively. 
The latency of the NN processing module is \(23.12  \mu s\) and the power consumption of the HD processing module is negligible compared to the NN processing module (i.e. less than 4\% of total power consumption).

\section{Related Work} \label{sec:related}

Kanerva \cite{kanerva2009hyperdimensional} explains the advantages and mathematical properties of HD computing, and how data patterns should correspond in a systematic way to the entities they represent in the real world for achieving brain-like computing.
Some of the prior work attempt to improve the performance of HD computing, either by increasing the obtained accuracy for some complex tasks, or enabling it to maintain the accuracy for lower dimensions. 
Authors in \cite{imani2018hierarchical} propose a hierarchical HD computing framework, which enables HD to improve its performance using multiple encoders without increasing the cost of classification.
In \cite{morris2019comphd}, authors utilize the mathematics of hyperdimensional spaces, and split each class hypervector into separate components and combine them into a reduced dimensional model.
However, these works have not explored the effect of the feature extraction for low-dimensional input features. 

Several studies in the literature explore hardware optimizations for implementing HD computing for different application domains. Authors in \cite{rahimi2016robust} propose a memory-centric architecture for the HD classifier with modular and scalable components, and demonstrate its performance on a language identification task. In \cite{datta2019programmable}, authors develop a programmable and scalable architecture for energy-efficient supervised classification using HD computing, and compare it with traditional architectures for a few conventional machine learning algorithms. The work in \cite{imani2017exploring} explores architectural designs for the cleanup memory to facilitate energy-efficient, fast, and scalable search operation, and the proposed designs are evaluated for a language recognition application.

\section{Conclusions}
\label{sec:conclusions}
In this paper, we proposed SynergicLearning, in which by designing NNs that include some components of the HD models in their training loop, we trained high-quality feature extraction layers tailored to the HD learning model. By passing the input low-dimensional features through these layers before encoding them into the HD space, the number of dimensions of the HD space was reduced by two to three orders of magnitude, while maintaining the high classification accuracy, which led to less complex HD classifier. We also proposed and implemented an end-to-end fully-parametrized implementation of SynergicLearning for inference. Following the proposed hardware architecture, we achieved 2.13x improvement in terms of latency, while yielding 1.60x lower power consumption compared to pure HD computing. 

\balance
\bibliographystyle{unsrt}
\bibliography{NNHDL}
	
\end{document}